\documentclass[a4paper,10pt]{article}

\usepackage{amsmath}
\usepackage{amsthm}
\usepackage{amssymb}
\usepackage{mathtools}
\usepackage{bm}

\usepackage{booktabs}

\usepackage{graphicx}
\usepackage{tango-colors}
\usepackage{tikz}
\usepackage{tikz-qtree}
\usetikzlibrary{arrows}
\usetikzlibrary{shapes}

\usepackage{pgfplots}
\pgfplotsset{compat=1.9}
\usepgfplotslibrary{groupplots}

\usepackage{algorithm}
\usepackage{algpseudocode}

\usepackage[backend=biber, style=authoryear]{biblatex}
\usepackage{hyperref}
\usepackage[english]{babel}
\usepackage[utf8]{inputenc}
\usepackage[T1]{fontenc}
\usepackage{csquotes}

\newtheorem{thm}{Theorem}

\theoremstyle{definition}

\newcommand{\mat}[1]{\bm{#1}}
\newcommand{\transp}[1]{{#1}^T}
\newcommand{\effic}{\mathcal{O}}

\newcommand{\grad}[1]{\nabla_{#1}}
\newcommand{\deriv}[1]{\frac{\partial}{\partial #1}}

\newcommand{\node}{x}
\newcommand{\lnode}{x}
\newcommand{\rnode}{y}

\newcommand{\ltree}{\bar x}
\newcommand{\rtree}{\bar y}

\newcommand{\trees}{\mathcal{T}}

\newcommand{\siz}[1]{|#1|}

\newcommand{\alphabet}{\mathcal{X}}
\newcommand{\sym}[1]{\text{\texttt{#1}}}
\newcommand{\gap}{-}

\newcommand{\pospairs}{P}
\newcommand{\negpairs}{N}
\newcommand{\loss}{E}

\newcommand{\margin}{\eta}
\newcommand{\regul}{\beta}

\newcommand{\alphidx}{u}
\newcommand{\ralphidx}{v}
\newcommand{\alphlim}{U}
\newcommand{\embedmat}{\mat A}
\newcommand{\embeddim}{V}
\newcommand{\embedvec}{\vec a}
\newcommand{\simplexfun}{\rho}

\newcommand{\relproj}{\Omega}

\newcommand{\script}{\bar \delta}

\newcommand{\dist}{d}
\newcommand{\cost}{c}

\newcommand{\freqmat}{\mat P}

\tikzstyle{point}=[circle, inner sep=0pt, minimum size=3mm, line width=0.5mm, anchor=center]
\tikzstyle{textnode}=[draw=none, fill=none]
\tikzstyle{proto}=[diamond, inner sep=0pt, minimum size=5mm, line width=0.5mm, anchor=center]
\tikzstyle{edge}=[->, >=stealth', shorten <=2pt, shorten >=2pt, auto, line width=0.5mm]
\tikzstyle{class0color}=[aluminium6]
\tikzstyle{class0}=[draw=aluminium6, fill=aluminium4, text=aluminium6]
\tikzstyle{class1color}=[skyblue3]
\tikzstyle{class1}=[draw=skyblue3, fill=skyblue1, text=skyblue3]
\tikzstyle{class2color}=[orange3]
\tikzstyle{class2}=[draw=orange3, fill=orange1, text=orange3]

\title{{\large Supplementary Materials and Results for the ICML 2018 paper}\\Tree Edit Distance Learning via Adaptive Symbol Embeddings}
\author{Benjamin Paaßen}
\date{2018-05-16}

\begin{document}

\maketitle

\section{The Tree Edit Distance Can be Computed Via Dynamic Programming}

\begin{thm}
Let $\cost$ be a pseudo-metric on $\alphabet \cup \{\gap\}$. Then, the
corresponding tree edit distance $\dist_\cost(\ltree, \rtree)$
can be computed in $\effic(\siz{\ltree}^2 \cdot \siz{\rtree}^2)$ 
using a dynamic programming scheme.

Conversely, if $\cost$ violates the triangular inequality, the dynamic
programming scheme overestimates the tree edit distance.

\begin{proof}
Refer to \textcite{Zhang1989} for a proof of the first claim. With respect to
the second claim, consider a three-letter alphabet $\alphabet = \{ a, b, c \}$
and a cost function which violates the triangular inequality with
$\cost(a, b) > \cost(a, c) + \cost(c, b) \geq 0$, but otherwise adheres to the triangular
inequality. In that case, the tree edit distance between $a$ and $b$ would be
$\cost(a, c) + \cost(c, b)$, because replacing $a$ with $c$ and $c$ with $b$ is
a valid edit script. However, this edit script is not considered by the recurrence
equations of \textcite{Zhang1989}, which only conside the options
$\dist_\cost(a, b) = \min \{ \cost(a, \gap) + \cost(\gap, b), \cost(a, b) \}$.
We can also extend this example to larger trees, for example by considering the
trees $a(c)$ and $b(c)$, which also would have the tree edit distance
$\cost(a, c) + \cost(c, b)$, and so on.
\end{proof}

\end{thm}

\section{A Pseudo-Metric Cost Function Implies A Pseudo-Metric Tree Edit Distance}

\begin{thm}
Let $\cost$ be a pseudo-metric on $\alphabet \cup \{\gap\}$. Then, the
corresponding tree edit distance $\dist_\cost$ is a pseudo-metric on the
set of possible trees over $\alphabet$.

However, if $\cost$ violates any of the pseudo-metric properties
(except for the triangular inequality), we can construct examples
such that $\dist_\cost$ violates the same pseudo-metric properties.

\begin{proof}
For the first part of the proof, refer to Theorem~1 in \textcite{Paassen2018arxiv}.

Further, consider the following counter-examples:
\begin{description}
\item[Non-negativity:] If $\cost$ breaks the non-negativity condition, this has the most
severe consequences for the edit distance, as the edit distance becomes ill-defined in many
cases. For example, let $\alphabet$ any alphabet and $\lnode, \rnode \in \alphabet$, such that
$\cost(\lnode, \rnode) < 0$. If $\cost$ is symmetric, we also have $\cost(\rnode, \lnode) < 0$.
However, this implies that any edit distance can be decreased arbitrarily. In particular, let
$\ltree, \rtree \in \trees(\alphabet)$ and let $\script$ be an edit script which transforms
$\ltree$ to $\rtree$. Then, we can prolong this edit script via an insertion of $\lnode$,
arbitrarily many replacements of $\lnode$ with $\rnode$, each followed by a replacement of
$\rnode$ with $\lnode$, and a deletion of $\lnode$. This way, we can reduce the cost of the
edit script arbitrarily, but maintain a valid script which transforms $\ltree$ to $\rtree$. Thus,
the edit distance becomes ill-defined (and certainly negative).
\item[Self-identity:] If $\cost$ breaks the self-identity condition, $\dist_\cost$ will also not
be self-identical, provided that $\cost$ maintains the triangular inequality. In particular,
let $\alphabet$ be any alphabet and let $\lnode \in \alphabet$, such that $\cost(\lnode, \lnode) > 0$.
Then, $\dist_\cost(\lnode, \lnode) = \cost(\lnode, \lnode) > 0$, because for any
$\rnode \in \alphabet$ it holds $\cost(\lnode, \rnode) + \cost(\rnode, \lnode)
\geq \cost(\lnode, \lnode)$ due to the triangular inequality.
\item[Symmetry:] If $\cost$ breaks the symmetry condition, $\dist_\cost$ will also be asymmetric,
provided that $\cost$ maintains the triangular inequality. In particular, let $\alphabet$ be any
alphabet and let $\lnode, \rnode \in \alphabet$, such that $\cost(\lnode, \rnode) < \cost(\rnode, \lnode)$.
Then it holds $\dist_\cost(\lnode, \rnode) \leq \cost(\lnode, \rnode) < \cost(\rnode, \lnode) = \dist_\cost(\rnode, \lnode)$.
The last equality holds because there can be no cheaper edit script between $\rnode$ and $\lnode$
than replacing $\rnode$ with $\lnode$ directly due to the triangular inequality.
\end{description}

\end{proof}
\end{thm}

\section{Good Edit Similarity Learning can Worsen The Edit Distance Loss}

\begin{thm}
There exists combinations of an alphabet $\alphabet$, positive pairs $\pospairs$, negative pairs
$\negpairs$, a default cost function $\cost_0$, and a regularization constant $\regul$, such that
the cost function $\cost_1$ learned by GESL is not a pseudo-metric, and yields a loss 
$\loss(\dist_{\cost_1}, \pospairs, \negpairs) > \loss(\tilde \dist_{\cost_1}, \pospairs, \negpairs)$,
as well as $\loss(\dist_{\cost_1}, \pospairs, \negpairs) > \loss(\dist_{\cost_0}, \pospairs, \negpairs)$.

\begin{proof}
We construct our example as follows: 
Let $\alphabet = \{ \sym{1}, \sym{2}, \sym{3} \}$ and consider the trees
$\ltree_1 = \sym{1}(\sym{2})$, $\ltree_2 = \sym{2}$, $\ltree_3 = \sym{3}$, and
$\ltree_4 = \sym{3}$ as well as the positive pairs $\pospairs = \{ (\ltree_1, \ltree_2), (\ltree_2, \ltree_1), (\ltree_3, \ltree_4), (\ltree_4, \ltree_3) \}$
and the negative pairs $\negpairs = \{ (\ltree_1, \ltree_3), (\ltree_2, \ltree_3), (\ltree_3, \ltree_1), (\ltree_4, \ltree_1) \}$,
and the default cost function $\cost_0(\lnode, \rnode) = \log(2)$ if $\lnode \neq \rnode$ and $0$ otherwise.
We further assume $0 < \regul < 1 / (5 \cdot \log(2))$, which appears as a reasonable range
for a regularization constant.

Note that the positive and negative pairs chosen here conform to the recommendation of \textcite{Bellet2012}
of selecting the closest neighbor of the same class and the furthest tree from a different class
for metric learning, if $\ltree_1$ and $\ltree_2$ are in class $1$ and $\ltree_3$ and $\ltree_4$
are in class $2$.

Our GESL optimization problem becomes:
\begin{align*}
\min_{\cost, \margin} \quad & \regul \cdot \lVert \cost \rVert^2 + [\tilde \dist_\cost(\ltree_1, \ltree_2) - \margin]_+ + [\log(2) + \margin - \tilde \dist_\cost(\ltree_1, \ltree_3)]_+ \\
+ & [\tilde \dist_\cost(\ltree_2, \ltree_1) - \margin]_+ + [\log(2) + \margin - \tilde \dist_\cost(\ltree_2, \ltree_3)]_+ \\
+ & [\tilde \dist_\cost(\ltree_3, \ltree_4) - \margin]_+ + [\log(2) + \margin - \tilde \dist_\cost(\ltree_3, \ltree_1)]_+ \\
+ & [\tilde \dist_\cost(\ltree_4, \ltree_3) - \margin]_+ + [\log(2) + \margin - \tilde \dist_\cost(\ltree_4, \ltree_1)]_+ \\
\text{s.t.} \quad & \forall \lnode, \rnode \in \alphabet : \cost(\lnode, \rnode) \geq 0, \quad \log(2) \geq \margin \geq 0
\end{align*}
Note that this minimization problem corresponds to setting the cost function values such that all
hinge loss terms become as close to zero as possible and, at the same time, all cost function values
are as small as possible.

\begin{table}
\caption{The initial cost function $\cost_0$ (top), and the learned cost functions
$\cost_1$ / $\cost_2$ via good edit similarity learning for the input trees
$\ltree_1 = \sym{1}(\sym{2})$ as well as $\ltree_2 = \sym{2}$, $\ltree_3 = \sym{3}$
in the first class, and $\ltree_3 = \sym{3}$ as well as
$\ltree_4 = \sym{3}$ in the second class. The symbol $\lnode$ that is replaced is shown
on the y axis, the symbol $\rnode$ it is replaced with is shown on the x axis.
The middle table refers to the learned cost function via standard GESL, the bottom
table refers to the learned cost function via GESL if \emph{all} cheapest edit scripts
are considered and metric properties are enforced.}
\label{tab:cost_functions}

\begin{center}
\begin{tabular}{ccccc}
$\cost_0(\lnode, \rnode)$   & $\sym{1}$ & $\sym{2}$ & $\sym{3}$ & $\gap$ \\
\cmidrule(lr){1-1} \cmidrule(lr){2-4} \cmidrule(lr){5-5}
$\sym{1}$ & $0$             & $\log(2)$       & $\log(2)$       & $\log(2)$    \\
$\sym{2}$ & $\log(2)$       & $0$             & $\log(2)$       & $\log(2)$    \\
$\sym{3}$ & $\log(2)$       & $\log(2)$       & $0$             & $\log(2)$    \\[0.2cm]
$\gap$    & $\log(2)$       & $\log(2)$       & $\log(2)$       & $0$
\end{tabular}

\vspace{0.3cm}

\begin{tabular}{ccccc}
$\cost_1(\lnode, \rnode)$   & $\sym{1}$     & $\sym{2}$    & $\sym{3}$     & $\gap$ \\
\cmidrule(lr){1-1} \cmidrule(lr){2-4} \cmidrule(lr){5-5}
$\sym{1}$ & $0$           & $0$          & $\frac{1}{2} \log(2)$ & $0$ \\
$\sym{2}$ & $0$           & $0$          & $\log(2)$     & $\frac{1}{2} \log(2)$ \\
$\sym{3}$ & $\frac{1}{2} \log(2)$ & $0$  & $0$           & $0$ \\[0.2cm]
$\gap$    & $0$  & $\frac{1}{2} \log(2)$ & $0$           & $0$
\end{tabular}

\vspace{0.3cm}

\begin{tabular}{ccccc}
$\cost_2(\lnode, \rnode)$   & $\sym{1}$     & $\sym{2}$    & $\sym{3}$     & $\gap$ \\
\cmidrule(lr){1-1} \cmidrule(lr){2-4} \cmidrule(lr){5-5}
$\sym{1}$ & $0$                   & $0$          & $\frac{1}{2} \log(2)$ & $0$ \\
$\sym{2}$ & $0$                   & $0$          & $\log(2)$ & $\frac{1}{2} \log(2)$ \\
$\sym{3}$ & $\frac{1}{2} \log(2)$ & $\log(2)$    & $0$       & $\frac{1}{2} \log(2)$ \\[0.2cm]
$\gap$    & $0$                   & $\frac{1}{2} \log(2)$ & $\frac{1}{2} \log(2)$ & $0$
\end{tabular}
\end{center}
\end{table}

\begin{figure}
\begin{center}
\begin{tikzpicture}[xscale=3, yscale=2.5]

\node[fill=none] (ab) at (0, 0) {$\ltree_1 = \sym{1}(\sym{2})$};
\node[fill=none] (b) at (0, -2) {$\ltree_1 = \sym{2}$};
\node[fill=none] (c) at (2, -1) {$\ltree_3 = \ltree_4 = \sym{3}$};

\path[edge, semithick]%
(ab) edge[bend left=15] node[rotate=90, above] {$(\sym{1}, \gap)$, $(\sym{2}, \sym{2})$} (b)
(ab) edge[densely dashed, bend left=45] node[rotate=-30, above] {$(\sym{1}, \gap)$, $(\sym{2}, \sym{3})$} (c)
(ab) edge[bend left=15] node[rotate=-30, above] {$(\sym{1}, \sym{3})$, $(\sym{2}, \gap)$} (c)
(b) edge[bend left=15] node[rotate=90, above] {$(\gap, \sym{1})$, $(\sym{2}, \sym{2})$} (ab)
(b) edge[bend left=15] node[rotate=25, above] {$(\sym{2}, \sym{3})$} (c)
(c) edge[bend left=15] node[rotate=-30, above] {$(\sym{3}, \sym{1})$, $(\gap, \sym{2})$} (ab)
(c) edge[densely dashed, bend left=45] node[rotate=-30, above] {$(\gap, \sym{1})$, $(\sym{3}, \sym{2})$} (ab)
(c) edge[bend left=15] node[rotate=25, above] {$(\sym{3}, \sym{2})$} (b)
(c) edge[loop right] node[right] {$(\sym{3}, \sym{3})$} (c);

\end{tikzpicture}
\end{center}
\caption{All co-optimal edit scripts according to the default cost function shown in Table~\ref{tab:cost_functions}, top,
between the trees $\ltree_1 = \sym{1}(\sym{2})$, $\ltree_2 = \sym{2}$, and $\ltree_3 = \sym{3}$.
Dashed edges signify co-optimal edit scripts which are not considered by GESL.}
\label{fig:edit_scripts}
\end{figure}
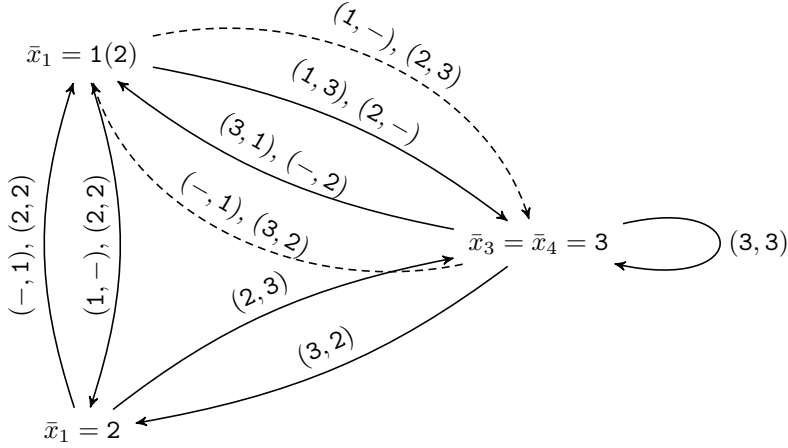

For the pseudo edit distances we obtain
\begin{align*}
\tilde \dist_\cost(\ltree_1, \ltree_2) = \cost(\sym{1}, \gap) + \cost(\sym{2}, \sym{2}), \quad &
\tilde \dist_\cost(\ltree_1, \ltree_3) = \cost(\sym{1}, \sym{3}) + \cost(\sym{2}, \gap) \\
\tilde \dist_\cost(\ltree_2, \ltree_1) = \cost(\gap, \sym{1}) + \cost(\sym{2}, \sym{2}), \quad &
\tilde \dist_\cost(\ltree_2, \ltree_3) = \cost(\sym{2}, \sym{3}) \\
\tilde \dist_\cost(\ltree_3, \ltree_4) = \cost(\sym{3}, \sym{3}), \quad &
\tilde \dist_\cost(\ltree_3, \ltree_1) = \cost(\sym{3}, \sym{1}) + \cost(\gap, \sym{2}) \\
\tilde \dist_\cost(\ltree_4, \ltree_3) = \cost(\sym{3}, \sym{3}), \quad &
\tilde \dist_\cost(\ltree_4, \ltree_1) = \cost(\sym{3}, \sym{1}) + \cost(\gap, \sym{2})
\end{align*}
based on the scripts shown in figure~\ref{fig:edit_scripts}, which yields the actual
quadratic problem:

\begin{align*}
\min_{\cost, \margin} \quad & \regul \cdot \lVert \cost \rVert^2
+   [\cost(\sym{1}, \gap) + \cost(\sym{2}, \sym{2}) - \margin]_+ + [\log(2) + \margin - \cost(\sym{1}, \sym{3}) - \cost(\sym{2}, \gap)]_+ \\
+ & [\cost(\gap, \sym{1}) + \cost(\sym{2}, \sym{2}) - \margin]_+ + [\log(2) + \margin - \cost(\sym{2}, \sym{3})]_+ \\
+ & 2 \cdot [\cost(\sym{3}, \sym{3}) - \margin]_+ + 2 \cdot [\log(2) + \margin - \cost(\sym{3}, \sym{1}) - \cost(\gap, \sym{2})]_+ \\
\text{s.t.} \quad & \forall \lnode, \rnode \in \alphabet : \cost(\lnode, \rnode) \geq 0, \quad \log(2) \geq \margin \geq 0
\end{align*}

To solve this problem, it is obviously optimal to set $\cost(\sym{1}, \sym{1}) =$ $\cost(\sym{2}, \sym{2}) =$
$\cost(\sym{3}, \sym{3}) =$ $\cost(\sym{1}, \sym{2}) =$ $\cost(\sym{2}, \sym{1}) =$ $\cost(\sym{1}, \gap) =$ $\cost(\gap, \sym{1}) =$ $\cost(\gap, \sym{3}) =$ $\cost(\sym{3}, \gap) =0$
as the cost function can be reduced with all these settings, without any influence on remaining settings.
Note that this also permits us to set $\margin = 0$, which reduces the hinge-loss contributions
for negative pairs. For $\cost(2, 3)$ consider the derivative $\deriv{\cost(2, 3)} (\regul \cdot \lVert \cost \rVert^2 + [\log(2) + \margin - \cost(\sym{2}, \sym{3})]_+)$,
which is $\regul \cdot 2 \cdot \cost(2, 3) - 1$ for $\cost(2, 3) \leq \log(2)$ and $\regul \cdot 2 \cdot \cost(2, 3) > 0$ for $\cost(2, 3) > \log(2)$. Because we assumed
$0 \leq \regul < 1 / (5 \cdot \log(2))$ this derivative is negative for $\cost(2, 3) \leq \log(2)$,
which means that we obtain an optimal value by setting $\cost(2, 3) = \log(2)$.
Using a similar reasoning, we obtain $\cost(\sym{1}, \sym{3}) = \cost(\sym{2}, \gap) = \cost(\sym{3}, \sym{1}) = \cost(\gap, \sym{2}) =
\log(2) / 2$, which ensures that all hinge-loss contributions are zero. We call this learned
cost function $\cost_1$.
For the final loss we obtain $\loss(\tilde \dist_{\cost_1}, \pospairs, \negpairs) = 4 \cdot \regul \cdot (\log(2) / 2)^2 + \regul \cdot \log(2)^2$.
The learned parameters are also listed in Table~\ref{tab:cost_functions}, middle.

First, notice that the learned cost function $\cost_1$ is not symmetric because
$0 = \cost_1(\sym{3}, \sym{2}) < \cost_1(\sym{2}, \sym{3}) = \log(2)$. Further,
$\cost_1$ also violates the triangular inequality because $\log(2) / 2 = \cost_1(\sym{2}, \sym{1}) + \cost_1(\sym{1}, \sym{3}) < \cost_1(\sym{2}, \sym{3}) = \log(2)$.

Furthermore, this loss considerably underestimates the loss $\loss(\dist_{\cost_1}, \pospairs, \negpairs)$.
This is due to the fact that the learned cost function permits to edit any tree into any other
without any cost, as is illustrated in figure~\ref{fig:degenerate_edit_scripts}.

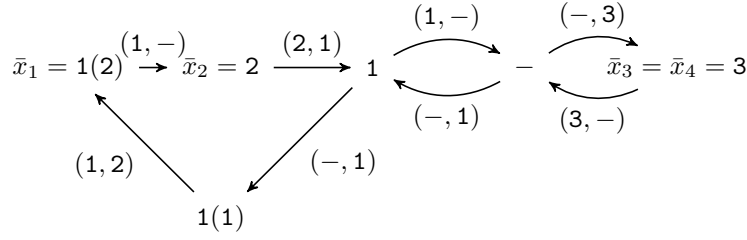
\begin{figure}
\begin{center}
\begin{tikzpicture}[xscale=2, yscale=2]

\node[fill=none] (ab) at (0, 0) {$\ltree_1 = \sym{1}(\sym{2})$};
\node[fill=none] (b)  at (1, 0) {$\ltree_2 = \sym{2}$};
\node[fill=none] (a)  at (2, 0) {$\sym{1}$};
\node[fill=none] (g)  at (3, 0) {$\gap$};
\node[fill=none] (c)  at (4, 0) {$\ltree_3 = \ltree_4 = \sym{3}$};
\node[fill=none] (aa) at (1,-1) {$\sym{1}(\sym{1})$};

\path[edge, semithick]%
(ab) edge node[above] {$(\sym{1}, \gap)$} (b)
(b)  edge node[above] {$(\sym{2}, \sym{1})$} (a)
(a)  edge[bend left] node[above] {$(\sym{1}, \gap)$} (g)
(g)  edge[bend left] node[above] {$(\gap, \sym{3})$} (c)
(c)  edge[bend left] node[below] {$(\sym{3}, \gap)$} (g)
(g)  edge[bend left] node[below] {$(\gap, \sym{1})$} (a)
(a)  edge node[below right] {$(\gap, \sym{1})$} (aa)
(aa) edge node[below left] {$(\sym{1}, \sym{2})$} (ab);

\end{tikzpicture}
\end{center}
\caption{A graph of intermediate trees where edges corresponds to edits with zero cost according to the
learned cost function $\cost_1$ obtained via GESL (see Table~\ref{tab:cost_functions}, middle).
Because we only consider zero-cost-edges, the edit distance between all connected trees in this
network degenerates to zero.}
\label{fig:degenerate_edit_scripts}
\end{figure}

Therefore, we obtain the loss $\loss(\dist_{\cost_1}, \pospairs, \negpairs) = 4 \cdot \regul \cdot (\log(2) / 2)^2 + \regul \cdot \log(2)^2 + 4 \cdot \log(2) = 2 \cdot \regul \cdot \log(2)^2 + 4 \cdot \log(2)$,
which is considerably larger than $\loss(\tilde \dist_{\cost_1}, \pospairs, \negpairs)$.

Further, consider the default loss according to the cost function $\cost_0$ in Table~\ref{tab:cost_functions}, top, which is:
\begin{align*}
\loss(\dist_{\cost_0}, \pospairs, \negpairs) = &12 \cdot \regul \cdot \log(2)^2 +
[\log(2) - \margin]_+ + [\log(2) + \margin - 2 \cdot \log(2)]_+ \\
+ & [\log(2) - \margin]_+ + [\log(2) + \margin - \log(2)]_+ \\
+ & [0 - \margin]_+ + [\log(2) + \margin - 2 \cdot \log(2)]_+ \\
+ & [0 - \margin]_+ + [\log(2) + \margin - 2 \cdot \log(2)]_+
\end{align*}
This loss depends on the value for $\margin$ and is lower for higher values of $\margin$.
At most, the loss is $12 \cdot \regul \cdot \log(2)^2 + 2 \cdot \log(2)$ (for $\margin = 0$).
Because we assumed $\regul < 1 / (5 \cdot \log(2))$, this loss is lower than
$\loss(\dist_{\cost_1}, \pospairs, \negpairs)$.
\end{proof}
\end{thm}

\paragraph{Remark:}
The previous theorem does not hold if $\freqmat_{\cost_0}$ measures the average over \emph{all}
cheapest edit scripts instead of just one cheapest edit script and if pseudo-metric properties are
enforced. In this case, the pseudo edit distances are computed as:

\begin{align*}
\tilde \dist_\cost(\ltree_1, \ltree_2) = \cost(\sym{1}, \gap) + \cost(\sym{2}, \sym{2}), \quad &
\tilde \dist_\cost(\ltree_1, \ltree_3) = \frac{1}{2} \cdot [\cost(\sym{1}, \sym{3}) + \cost(\sym{2}, \gap)] + \frac{1}{2} \cdot [\cost(\sym{1}, \gap) + \cost(\sym{2}, \sym{3})]  \\
\tilde \dist_\cost(\ltree_2, \ltree_1) = \cost(\gap, \sym{1}) + \cost(\sym{2}, \sym{2}), \quad &
\tilde \dist_\cost(\ltree_2, \ltree_3) = \cost(\sym{2}, \sym{3}) \\
\tilde \dist_\cost(\ltree_3, \ltree_4) = \cost(\sym{3}, \sym{3}), \quad &
\tilde \dist_\cost(\ltree_3, \ltree_1) = \frac{1}{2} \cdot [\cost(\sym{3}, \sym{1}) + \cost(\gap, \sym{2})] + \frac{1}{2} \cdot [\cost(\gap, \sym{1}) + \cost(\sym{3}, \sym{2})] \\
\tilde \dist_\cost(\ltree_4, \ltree_3) = \cost(\sym{3}, \sym{3}), \quad &
\tilde \dist_\cost(\ltree_4, \ltree_1) = \frac{1}{2} \cdot [\cost(\sym{3}, \sym{1}) + \cost(\gap, \sym{2})] + \frac{1}{2} \cdot [\cost(\gap, \sym{1}) + \cost(\sym{3}, \sym{2})]
\end{align*}
yielding the optimization problem
\begin{align*}
\min_{\cost, \margin} \quad & \regul \cdot \lVert \cost \rVert^2
+   [\cost(\sym{1}, \gap) + \cost(\sym{2}, \sym{2}) - \margin]_+ + [\log(2) + \margin - \frac{1}{2} \cdot \big(\cost(\sym{1}, \sym{3}) + \cost(\sym{2}, \gap) + \cost(\sym{1}, \gap) + \cost(\sym{2}, \sym{3})\big)]_+ \\
+ & [\cost(\gap, \sym{1}) + \cost(\sym{2}, \sym{2}) - \margin]_+ + [\log(2) + \margin - \cost(\sym{2}, \sym{3})]_+ \\
+ & 2 \cdot [\cost(\sym{3}, \sym{3}) - \margin]_+ + 2 \cdot [\log(2) + \margin - \frac{1}{2} \cdot \big( \cost(\sym{3}, \sym{1}) + \cost(\gap, \sym{2}) + \cost(\gap, \sym{1}) + \cost(\sym{3}, \sym{2})\big)]_+ \\
\text{s.t.} \quad & \forall \lnode, \rnode \in \alphabet : \cost(\lnode, \rnode) \geq 0, \quad \log(2) \geq \margin \geq 0, \quad \cost \text{ is pseudo-metric}
\end{align*}

First, we observe that we can set $\cost(\sym{1}, \sym{1}) = \cost(\sym{2}, \sym{2}) =
\cost(\sym{3}, \sym{3}) = \cost(\sym{1}, \sym{2}) = \cost(\sym{2}, \sym{1}) = 0$
because the derivative of the cost function with respect to these parameters is positive at all values
$\geq 0$, independent of all other parameters.

Next, note that the parameters $\cost(\sym{1}, \gap)$ and $\cost(\gap, \sym{2})$ are coupled due to
the symmetry requirement. We obtain a derivative of $\deriv{\cost(\sym{1}, \gap)} (\regul \cdot 2 \cdot \cost(\sym{1}, \gap)^2$
$+   [\cost(\sym{1}, \gap) + \cost(\sym{2}, \sym{2}) - \margin]_+$
$+ [\log(2) + \margin - \frac{1}{2} \cdot \big(\cost(\sym{1}, \sym{3}) + \cost(\sym{2}, \gap) + \cost(\sym{1}, \gap) + \cost(\sym{2}, \sym{3})\big)]_+$
$+ [\cost(\gap, \sym{1}) + \cost(\sym{2}, \sym{2}) - \margin]_+$
$+ 2 \cdot [\log(2) + \margin - \frac{1}{2} \cdot \big( \cost(\sym{3}, \sym{1}) + \cost(\gap, \sym{2}) + \cost(\gap, \sym{1}) + \cost(\sym{3}, \sym{2})\big)]_+)$.
$= 4 \cdot \regul \cdot \cost(\sym{1}, \gap) + 1 - \frac{1}{2} + 1 - 2 \cdot \frac{1}{2}$ if the
hinge-loss contributions are positive, which means that an optimum lies at $\cost(\sym{1}, \gap) = \cost(\gap, \sym{1}) = \margin$,
depending on the value of $\margin$.

Next, note that the parameters $\cost(\sym{2}, \gap)$ and $\cost(\gap, \sym{2})$ are coupled due to
the symmetry requirement. We obtain a derivative of $\deriv{\cost(\sym{2}, \gap)} (\regul \cdot 2 \cdot \cost(\sym{2}, \gap)^2$
$+ [\log(2) + \margin - \frac{1}{2} \cdot \big(\cost(\sym{1}, \sym{3}) + \cost(\sym{2}, \gap) + \cost(\sym{1}, \gap) + \cost(\sym{2}, \sym{3})\big)]_+$
$+ 2 \cdot [\log(2) + \margin - \frac{1}{2} \cdot \big( \cost(\sym{3}, \sym{1}) + \cost(\gap, \sym{2}) + \cost(\gap, \sym{1}) + \cost(\sym{3}, \sym{2})\big)]_+)$
$= 4 \cdot \regul \cdot \cost(\sym{2}, \gap) - \frac{1}{2} - 2 \cdot \frac{1}{2}$ as long as
the hinge-loss contributions are non-zero. Because we assumed $\regul < 1 / (5 \cdot \log(2))$ this gradient is negative
for $\cost(\sym{2}, \gap) \leq \frac{15}{8} \log(2)$, dependent on the values of $\cost(\sym{1}, \sym{3}) = \cost(\sym{3}, \sym{1})$,
$\cost(\sym{2}, \sym{3}) = \cost(\sym{3}, \sym{2})$, and $\cost(\sym{1}, \gap) = \cost(\gap, \sym{1})$.
Note that we can apply a similar reasoning for the parameters $\cost(\sym{1}, \sym{3}) = \cost(\sym{3}, \sym{1})$
and $\cost(\sym{2}, \sym{3}) = \cost(\sym{3}, \sym{2})$. We obtain the strongest negative gradient for
$\cost(\sym{3}, \sym{2})$ because it is needed to obtain a zero value for the hinge-loss contribution
$[\log(2) + \margin - \cost(\sym{2}, \sym{3})]_+$. In particular, we can ensure a zero value of
this contribution for $\cost(\sym{2}, \sym{3}) = \log(2) + \margin$. This leaves the
requirement $\log(2) + \margin - \frac{1}{2} \cdot \big(\cost(\sym{1}, \sym{3}) + \cost(\sym{2}, \gap) + \cost(\sym{1}, \gap) + \cost(\sym{2}, \sym{3}) \Big) \leq 0$
to ensure zero hinge-loss contributions, which is equivalent to
$\log(2) + \cost(\gap, \sym{1}) - \frac{1}{2} \cdot \big(\cost(\sym{1}, \sym{3}) + \cost(\sym{2}, \gap) + \cost(\sym{1}, \gap) + \log(2) + \cost(\gap, \sym{1})\Big) \leq 0$
$\iff \frac{1}{2} \log(2) - \frac{1}{2} \cdot \cost(\sym{1}, \sym{3}) - \frac{1}{2} \cdot \cost(\sym{2}, \gap) \leq 0$,
which in combination with the regularization yields the optimum $\cost(\sym{1}, \sym{3}) = \cost(\sym{3}, \sym{1}) = \cost(\sym{2}, \gap) = \cost(\gap, \sym{2}) = \frac{1}{2} \log(2)$.

Finally, we need to find an optimum for $\margin$, which then determines the values of
$\cost(\sym{1}, \gap) = \margin$ and  $\cost(\sym{2}, \sym{3}) = \log(2) + \margin$.
Due to the triangular inequality requirement we need to ensure that $\frac{1}{2} \log(2) = \cost(\sym{1}, \sym{3})
\leq \cost(\sym{1}, \gap) + \cost(\gap, \sym{3})$ and $\log(2) + \margin = \cost(\sym{2}, \sym{3}) \leq \cost(\sym{2}, \gap) + \cost(\gap, \sym{3}) = \frac{1}{2} \log(2) + \cost(\gap, \sym{3})$.
From those two inreualities we obtain $\frac{1}{2} \log(2) \leq \margin + \cost(\gap, \sym{3})$ and
$\frac{1}{2} \log(2) \leq -\margin + \cost(\gap, \sym{3})$. Due to the regularization, we wish
to set $\cost(\gap, \sym{3})$ as small as possible, which implies $\margin = 0$ and $\cost(\gap, \sym{3}) = \frac{1}{2} \log(2)$.
The final cost function $\cost_2$ is displayed in Table~\ref{tab:cost_functions}, bottom.
We obtain the same loss for the pseudo-edit distance as for the edit distance in this case,
namely $\loss(\tilde \dist_{\cost_2}, \pospairs, \negpairs) = \loss(\dist_{\cost_2}, \pospairs, \negpairs)
= 6 \cdot \regul \cdot (\frac{1}{2} \log(2))^2 + 2 \cdot \regul \cdot \log(2)^2 = \frac{7}{2} \cdot \regul \cdot \log(2)^2 < \frac{7}{10} \log(2)$.

\section{Embedding Initialization}

As initialization of the vectorial embedding we use the $\alphlim$-dimensional simplex with
side length $1$, which leads to $\cost_{\embedmat}(\lnode, \rnode) = 0$ if $\lnode = \rnode$
and $1$ otherwise.

\begin{algorithm}
\caption{An algorithm to construct a $\alphlim$-dimensional simplex with side length $1$.}
\label{alg:simplex}
\begin{algorithmic}
\Function{simplex}{dimension $\alphlim$}
\State{Initialize $\embedmat \gets \mat{0}^{\alphlim \times \alphlim}$.}
\State{Initialize $\simplexfun \gets 0^\alphlim$.}
\For{$\alphidx \gets 1, \ldots, \alphlim$}
\For{$\ralphidx \gets 1, \ldots, \alphidx - 1$.}
\State $\embedmat_{\ralphidx, \alphidx} \gets \simplexfun_\ralphidx$.
\EndFor
\State $\simplexfun_\alphidx \gets 1 / \sqrt{2 \cdot \alphidx \cdot (\alphidx + 1)}$.
\State $\embedmat_{\alphidx, \alphidx} \gets \simplexfun_\alphidx \cdot (\alphidx + 1)$.
\EndFor
\State \Return $\embedmat$.
\EndFunction
\end{algorithmic}
\end{algorithm}

\begin{thm}
An $\alphlim$-dimensional simplex with side length $1$ can be constructed using
algorithm~\ref{alg:simplex}. More precisely: All columns of the matrix $\embedmat$ returned by
algorithm~\ref{alg:simplex} have norm $1$ and pairwise Euclidean distance $1$.

\begin{proof}
We first show that $\sqrt{1 - \sum_{\ralphidx = 1}^{\alphidx - 1} \simplexfun_\ralphidx^2} = \simplexfun_\alphidx \cdot (\alphidx + 1)$
via induction. We observe that $\sqrt{1 - \sum_{\ralphidx = 1}^{1 - 1} \simplexfun_\ralphidx^2} = \sqrt{1 - 0} = 1 = \frac{1}{2} \cdot 2 = \frac{1}{\sqrt{2 \cdot 1 \cdot (1 + 1)}} \cdot (1 + 1) = \simplexfun_1 \cdot (1 + 1)$.
Now, let's assume that the claim holds for all $\alphidx' \leq \alphidx$ and consider $\alphidx + 1$.
Then, we obtain
\begin{align*}
\sqrt{1 - \sum_{\ralphidx = 1}^\alphidx \simplexfun_\ralphidx^2}
&= \sqrt{\Big(1 - \sum_{\ralphidx = 1}^{\alphidx - 1} \simplexfun_\ralphidx^2\Big) - \simplexfun_\alphidx^2} \\
&\stackrel{I.H.}{=} \sqrt{\Big(\simplexfun_\alphidx \cdot (\alphidx+1)\Big)^2 - \simplexfun_\alphidx^2} \\
&= \simplexfun_\alphidx \cdot \sqrt{(\alphidx+1)^2 - 1} \\
&= \sqrt{\frac{(\alphidx+1)^2 - 1}{2 \cdot \alphidx \cdot (\alphidx + 1)}} \\
&= \sqrt{\frac{\alphidx^2 + 2 \cdot \alphidx}{2 \cdot \alphidx \cdot (\alphidx + 1)}} \\
&= \sqrt{\frac{\alphidx + 2}{2 \cdot (\alphidx + 1)}} \cdot \frac{\alphidx + 2}{\alphidx + 2}\\
&= \frac{\alphidx + 2}{\sqrt{2 \cdot (\alphidx + 1) \cdot (\alphidx + 2)}}\\
&= \simplexfun_{\alphidx+1} \cdot (\alphidx + 2)\\
\end{align*}
which completes the induction.

Next, consider the norm of any column $\embedmat_{-, \alphidx}$ of $\embedmat$. Here, we obtain:
\begin{align*}
\lVert \embedmat_{-, \alphidx} \rVert &= \sqrt{ \sum_{\ralphidx = 1}^\alphlim \embedmat_{\ralphidx, \alphidx}^2 }
= \sqrt{ \sum_{\ralphidx = 1}^{\alphidx - 1} \embedmat_{\ralphidx, \alphidx}^2 + \embedmat_{\alphidx, \alphidx}^2}
= \sqrt{ \sum_{\ralphidx = 1}^{\alphidx - 1} \simplexfun_\ralphidx^2 + (\simplexfun_\alphidx \cdot (\alphidx + 1))^2} \\
&= \sqrt{ \sum_{\ralphidx = 1}^{\alphidx - 1} \simplexfun_\ralphidx^2 + 1 - \sum_{\ralphidx = 1}^{\alphidx - 1} \simplexfun_\ralphidx^2}
= \sqrt{1} = 1
\end{align*}

Finally, consider the Euclidean distance between any two columns $\embedmat_{-, \alphidx}$ and
$\embedmat_{-, \ralphidx}$ of $\embedmat$. Without loss of generality, assume that $\ralphidx > \alphidx$.
Here, we obtain:
\begin{align*}
\lVert \embedmat_{-, \alphidx} - \embedmat_{-, \ralphidx} \rVert &=
\sqrt{ \sum_{\ralphidx' = 1}^\alphlim (\embedmat_{\ralphidx', \alphidx} - \embedmat_{\ralphidx', \ralphidx})^2 } \\
&= \sqrt{ \sum_{\ralphidx' = 1}^{\alphidx - 1} (\embedmat_{\ralphidx', \alphidx} - \embedmat_{\ralphidx', \ralphidx})^2
+ (\embedmat_{\alphidx, \alphidx} - \embedmat_{\alphidx, \ralphidx})^2
+ \sum_{\ralphidx' = \alphidx+1}^{\ralphidx - 1} (\embedmat_{\ralphidx', \alphidx} - \embedmat_{\ralphidx', \ralphidx})^2
+ (\embedmat_{\ralphidx, \alphidx} - \embedmat_{\ralphidx, \ralphidx})^2 } \\
&= \sqrt{ \sum_{\ralphidx' = 1}^{\alphidx - 1} (\simplexfun_{\ralphidx'} - \simplexfun_{\ralphidx'})^2
+ (\simplexfun_\alphidx \cdot (\alphidx + 1) - \simplexfun_\alphidx)^2
+ \sum_{\ralphidx' = \alphidx+1}^{\ralphidx - 1} (0 - \simplexfun_{\ralphidx'})^2
+ (0- \simplexfun_{\ralphidx, \ralphidx} \cdot (\ralphidx+1))^2 } \\
&= \sqrt{\simplexfun_\alphidx^2 \cdot (\alphidx + 1 - 1)^2
+ \sum_{\ralphidx' = \alphidx+1}^{\ralphidx - 1} (\simplexfun_{\ralphidx'})^2
+ 1 - \sum_{\ralphidx' = 1}^{\ralphidx - 1} (\simplexfun_{\ralphidx'})^2 } \\
&= \sqrt{\simplexfun_\alphidx^2 \cdot \alphidx^2
+ 1 - \sum_{\ralphidx' = 1}^\alphidx (\simplexfun_{\ralphidx'})^2 } \\
&= \sqrt{\simplexfun_\alphidx^2 \cdot \alphidx^2 + \simplexfun_{\alphidx + 1}^2 \cdot (\alphidx + 2)^2 } \\
&= \sqrt{\frac{\alphidx^2}{2 \cdot \alphidx \cdot (\alphidx + 1)}
+ \frac{(\alphidx+2)^2}{2 \cdot (\alphidx+1) \cdot (\alphidx + 2)} } \\
&= \sqrt{\frac{\alphidx}{2 \cdot (\alphidx + 1)}
+ \frac{\alphidx+2}{2 \cdot (\alphidx+1)} }
= \sqrt{\frac{\alphidx + \alphidx + 2}{2 \cdot (\alphidx + 1)} } = 1 \\
\end{align*}
which concludes the proof.
\end{proof}
\end{thm}

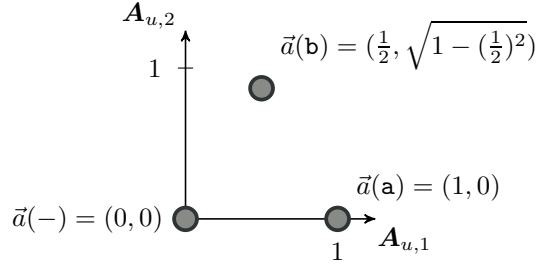
\begin{figure}
\begin{center}
\begin{tikzpicture}[scale=2]

\draw[->, >=stealth', semithick] (0, 0) -- (1.25, 0);
\node[below right] at (1.2, 0) {$\embedmat_{\alphidx, 1}$};
\draw[->, >=stealth', semithick] (0, 0) -- (0, 1.25);
\node[above left]  at (0, 1.2) {$\embedmat_{\alphidx, 2}$};


\draw (1, -0.05) -- (1, 0.05);
\node[below, outer sep=0.2cm] at (1,0) {$1$};
\draw (-0.05, 1) -- (0.05, 1);
\node[left, outer sep=0.2cm] at (0,1) {$1$};


\node[point, class0] (gap) at (0, 0) [label=left:{$\embedvec(\gap) = (0, 0)$}]    {};
\node[point, class0] (a)   at (1, 0) [label=above right:{$\embedvec(\sym{a}) = (1, 0)$}] {};
\node[point, class0] (b)   at (60:1) [label=above right:{$\embedvec(\sym{b}) = (\frac{1}{2}, \sqrt{1 - (\frac{1}{2})^2})$}] {};

\end{tikzpicture}
\end{center}
\caption{The initial 2-dimensional embedding of the alphabet $\alphabet = \{ \sym{a}, \sym{b} \}$.
This embedding ensures that the pairwise distance of all symbols, including the $\gap$-symbol, is $1$.}
\label{fig:simplex}
\end{figure}

An example initialization embedding for the alphabet $\alphabet = \{ \sym{a}, \sym{b}\}$
is shown in figure~\ref{fig:simplex}.

\section{Cosine Distance Gradient}

Consider the following cost function including the $\embeddim \times \embeddim$ matrix $\relproj$:
\begin{align}
\cost_\relproj(\vec \lnode, \vec \rnode) &= \frac{1}{2} \cdot \Big(1 - s_\relproj(\vec \lnode, \vec \rnode) \Big) & \text{where} \\
s_\relproj(\vec \lnode, \vec \rnode) &=
\frac{
\transp{(\relproj \cdot \vec \lnode)} \cdot \relproj \cdot \vec \rnode
}{
\lVert \relproj \cdot \vec \lnode \rVert \cdot \lVert \relproj \cdot \vec \rnode \rVert
}
\end{align}
The gradient of that function with respect to $\relproj$ is:
\begin{align}
&\grad{\relproj} \cost_\relproj(\vec \lnode, \vec \rnode) = -\frac{1}{2} \grad{\relproj}s_\relproj(\vec \lnode, \vec \rnode) \\
&=-
\frac{(\relproj \cdot \vec \lnode) \cdot \transp{\vec \rnode} + (\relproj \cdot \vec \rnode) \cdot \transp{\vec \lnode}
- s_\relproj(\vec \lnode, \vec \rnode) \cdot \Big[
(\relproj \cdot \vec \lnode) \cdot \transp{\vec \lnode} \cdot \frac{ \lVert \relproj \cdot \vec \rnode \rVert}{\lVert \relproj \cdot \vec \lnode \rVert} +
(\relproj \cdot \vec \rnode) \cdot \transp{\vec \rnode} \cdot \frac{ \lVert \relproj \cdot \vec \lnode \rVert}{\lVert \relproj \cdot \vec \rnode \rVert}
\Big]
}{
2 \cdot \lVert \relproj \cdot \vec \lnode \rVert \cdot \lVert \relproj \cdot \vec \rnode \rVert
} \notag
\end{align}

\section{Ablation Studies}

In the main paper, we display the average classification error on all data sets for
good edit similarity learning (GESL) using the pseudo edit distance and our proposed approach using the actual edit distance.
However, our approach differs from GESL in several design decisions, each of which may have
influence on the resulting classification accuracy. In Figure~\ref{fig:ablation} we display
the average classification error and standard deviation (as error bars) for all data sets
and the following different design options.
\begin{enumerate}
\item Classic GESL (G1),
\item GESL using the average over all co-optimal edit scripts for $\freqmat_{\cost_0}$, instead
of only one co-optimal edit script (G2),
\item GESL using the average over all co-optimal edit scripts and the median learning vector
quantization (G3),
\item LVQ metric learning, directly learning the cost function parameters instead of an embedding,
with a pseudo-metric normalization after each gradient step (L1), and
\item LVQ embedding metric learning as proposed in the paper (L2).
\end{enumerate}

We interpret these results as follows (quoted from the paper):
\enquote{We observed that considering the average over all co-optimal edit scripts,
and considering LVQ prototypes instead of ad-hoc nearest neighbors improved GESL
on the MiniPalindrome data set, worsened it for the strings data set, and otherwise
showed no remarkable difference for the Sorting, Cystic, and Leukemia data set.
We also compared LVQ metric learning without the embedding approach and with the
embedding approach. Interestingly, the pseudo-edit distance performed worse when
considering embeddings, while the actual edit distance performed better when
considering embeddings. In general, GESL variants performed better for the
pseudo-edit distance than for the actual edit distance, and LVQ variants performed
better for the actual edit dsitance compared to the pseudo edit distance.}

\pgfplotsset{
	errline/.style={%
		error bars/.cd,
		y dir=both,y explicit
	},
	knn/.style={skyblue3, mark=o, errline},
	lvq/.style={orange3, xshift=0.05cm, mark=square, errline},
	svm/.style={plum3, xshift=0.1cm, mark=triangle, mark size=0.9mm, errline},
	good/.style={scarletred3, xshift=0.15cm, mark=diamond, mark size=0.9mm, errline},
}

\begin{figure}
\begin{tikzpicture}
\begin{groupplot}[
	group style={
		group size=2 by 5,
		horizontal sep=0.1cm,
		vertical sep=1cm,
		x descriptions at=edge bottom,
		y descriptions at=edge left,
	},
	height=4cm,
	width=6.5cm,
	xtick={1,2,3,4,5},
	xticklabels={G1, G2, G3, L1, L2},
	xmin=1,xmax=5,
	enlarge x limits=0.1,
	ymin=0,
	ymax=0.5,
	ytick={0,0.1,0.2,0.3,0.4,0.5},
	enlarge y limits=0.1,
	legend cell align=left,
	legend pos=outer north east,
]
\nextgroupplot[title={\hspace{5.5cm}Strings}]
\addplot[aluminium3, sharp plot, ultra thin, update limits=false, forget plot]
coordinates {(0.001,0) (5,0)};
\addplot[knn] table[x=m, y=knn_mean, y error=knn_std] {string_experiment_results_pseudo-edit_distance.csv};
\addplot[lvq] table[x=m, y=mrglvq_mean, y error=mrglvq_std] {string_experiment_results_pseudo-edit_distance.csv};
\addplot[svm] table[x=m, y=svm_mean, y error=svm_std] {string_experiment_results_pseudo-edit_distance.csv};
\addplot[good] table[x=m, y=good_mean, y error=good_std] {string_experiment_results_pseudo-edit_distance.csv};
\nextgroupplot
\addplot[aluminium3, sharp plot, ultra thin, update limits=false, forget plot]
coordinates {(0.001,0) (5,0)};
\addplot[knn] table[x=m, y=knn_mean, y error=knn_std] {string_experiment_results_edit_distance.csv};
\addlegendentry{KNN}
\addplot[lvq] table[x=m, y=mrglvq_mean, y error=mrglvq_std] {string_experiment_results_edit_distance.csv};
\addlegendentry{MRGLVQ}
\addplot[svm] table[x=m, y=svm_mean, y error=svm_std] {string_experiment_results_edit_distance.csv};
\addlegendentry{SVM}
\addplot[good] table[x=m, y=good_mean, y error=good_std] {string_experiment_results_edit_distance.csv};
\addlegendentry{goodness}
\nextgroupplot[title={\hspace{5.5cm}MiniPalindrome}]
\addplot[aluminium3, sharp plot, ultra thin, update limits=false, forget plot]
coordinates {(0.001,0) (5,0)};
\addplot[knn] table[x=m, y=knn_mean, y error=knn_std] {mini_palindrome_experiment_results_pseudo-edit_distance.csv};
\addplot[lvq] table[x=m, y=mrglvq_mean, y error=mrglvq_std] {mini_palindrome_experiment_results_pseudo-edit_distance.csv};
\addplot[svm] table[x=m, y=svm_mean, y error=svm_std] {mini_palindrome_experiment_results_pseudo-edit_distance.csv};
\addplot[good] table[x=m, y=good_mean, y error=good_std] {mini_palindrome_experiment_results_pseudo-edit_distance.csv};
\nextgroupplot
\addplot[aluminium3, sharp plot, ultra thin, update limits=false, forget plot]
coordinates {(0.001,0) (5,0)};
\addplot[knn] table[x=m, y=knn_mean, y error=knn_std] {mini_palindrome_experiment_results_edit_distance.csv};
\addplot[lvq] table[x=m, y=mrglvq_mean, y error=mrglvq_std] {mini_palindrome_experiment_results_edit_distance.csv};
\addplot[svm] table[x=m, y=svm_mean, y error=svm_std] {mini_palindrome_experiment_results_edit_distance.csv};
\addplot[good] table[x=m, y=good_mean, y error=good_std] {mini_palindrome_experiment_results_edit_distance.csv};
\nextgroupplot[title={\hspace{5.5cm}Sorting}]
\addplot[aluminium3, sharp plot, ultra thin, update limits=false, forget plot]
coordinates {(0.001,0) (5,0)};
\addplot[knn] table[x=m, y=knn_mean, y error=knn_std] {sorting_experiment_results_pseudo-edit_distance.csv};
\addplot[lvq] table[x=m, y=mrglvq_mean, y error=mrglvq_std] {sorting_experiment_results_pseudo-edit_distance.csv};
\addplot[svm] table[x=m, y=svm_mean, y error=svm_std] {sorting_experiment_results_pseudo-edit_distance.csv};
\addplot[good] table[x=m, y=good_mean, y error=good_std] {sorting_experiment_results_pseudo-edit_distance.csv};
\nextgroupplot
\addplot[aluminium3, sharp plot, ultra thin, update limits=false, forget plot]
coordinates {(0.001,0) (5,0)};
\addplot[knn] table[x=m, y=knn_mean, y error=knn_std] {sorting_experiment_results_edit_distance.csv};
\addplot[lvq] table[x=m, y=mrglvq_mean, y error=mrglvq_std] {sorting_experiment_results_edit_distance.csv};
\addplot[svm] table[x=m, y=svm_mean, y error=svm_std] {sorting_experiment_results_edit_distance.csv};
\addplot[good] table[x=m, y=good_mean, y error=good_std] {sorting_experiment_results_edit_distance.csv};
\nextgroupplot[title={\hspace{5.5cm}Cystic}]
\addplot[aluminium3, sharp plot, ultra thin, update limits=false, forget plot]
coordinates {(0.001,0) (5,0)};
\addplot[knn] table[x=m, y=knn_mean, y error=knn_std] {cystic_experiment_results_pseudo-edit_distance.csv};
\addplot[lvq] table[x=m, y=mrglvq_mean, y error=mrglvq_std] {cystic_experiment_results_pseudo-edit_distance.csv};
\addplot[svm] table[x=m, y=svm_mean, y error=svm_std] {cystic_experiment_results_pseudo-edit_distance.csv};
\addplot[good] table[x=m, y=good_mean, y error=good_std] {cystic_experiment_results_pseudo-edit_distance.csv};
\nextgroupplot
\addplot[aluminium3, sharp plot, ultra thin, update limits=false, forget plot]
coordinates {(0.001,0) (5,0)};
\addplot[knn] table[x=m, y=knn_mean, y error=knn_std] {cystic_experiment_results_edit_distance.csv};
\addplot[lvq] table[x=m, y=mrglvq_mean, y error=mrglvq_std] {cystic_experiment_results_edit_distance.csv};
\addplot[svm] table[x=m, y=svm_mean, y error=svm_std] {cystic_experiment_results_edit_distance.csv};
\addplot[good] table[x=m, y=good_mean, y error=good_std] {cystic_experiment_results_edit_distance.csv};
\nextgroupplot[title={\hspace{5.5cm}Leukemia}]
\addplot[aluminium3, sharp plot, ultra thin, update limits=false, forget plot]
coordinates {(0.001,0) (5,0)};
\addplot[knn] table[x=m, y=knn_mean, y error=knn_std] {leukemia_experiment_results_pseudo-edit_distance.csv};
\addplot[lvq] table[x=m, y=mrglvq_mean, y error=mrglvq_std] {leukemia_experiment_results_pseudo-edit_distance.csv};
\addplot[svm] table[x=m, y=svm_mean, y error=svm_std] {leukemia_experiment_results_pseudo-edit_distance.csv};
\addplot[good] table[x=m, y=good_mean, y error=good_std] {leukemia_experiment_results_pseudo-edit_distance.csv};
\nextgroupplot
\addplot[aluminium3, sharp plot, ultra thin, update limits=false, forget plot]
coordinates {(0.001,0) (5,0)};
\addplot[knn] table[x=m, y=knn_mean, y error=knn_std] {leukemia_experiment_results_edit_distance.csv};
\addplot[lvq] table[x=m, y=mrglvq_mean, y error=mrglvq_std] {leukemia_experiment_results_edit_distance.csv};
\addplot[svm] table[x=m, y=svm_mean, y error=svm_std] {leukemia_experiment_results_edit_distance.csv};
\addplot[good] table[x=m, y=good_mean, y error=good_std] {leukemia_experiment_results_edit_distance.csv};
\end{groupplot}
\end{tikzpicture}
\caption{Ablation results for all data sets. Each row of the figure shows the results for one data set.
The left column shows the results for the pseudo-edit distance, the right column for the actual
edit distance. The x-axis in each plot displays the different design choices as described in the text
(from G1 to L2), the y-axis in each plot displays the mean classification accuracy after metric learning,
averaged across crossvalidation trials, with error bars displaying the standard deviation. The different
lines in each plot display the different classifiers used for evaluation.}
\label{fig:ablation}
\end{figure}
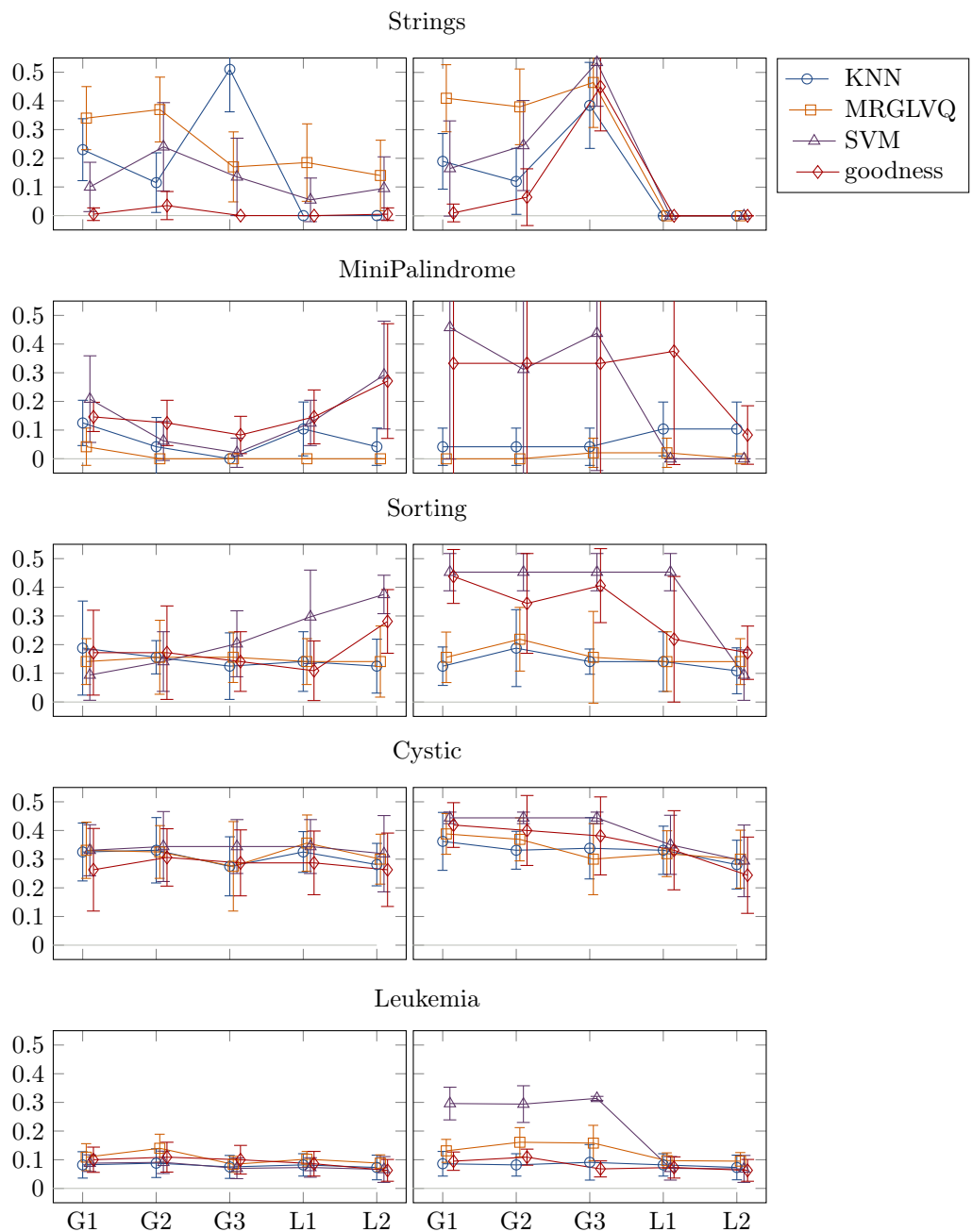

\printbibliography

\end{document}